\documentclass[]{article}
\usepackage[tagged, highstructure]{accessibility}
\usepackage{todonotes}
\usepackage{graphicx}
\usepackage{authblk}
\usepackage{subcaption}
\usepackage[hyphens]{url}
\usepackage{hyperref}
\hypersetup{breaklinks=true}
\usepackage{cite}
\usepackage{multirow}
\begin{document}

\title{A Pre-study on Data Processing Pipelines for Roadside Object Detection Systems Towards Safer Road Infrastructure}
\author[1,5]{Yinan Yu}
\author[1]{Samuel Scheidegger}
\author[2]{John-Fredrik Gr\"{o}nvall}
\author[3]{Magnus Palm}
\author[2]{Erik Svanberg}
\author[4]{Johan Amoruso Wennerby}
\author[1]{J\"{o}rg Bakker}
\affil[1]{Asymptotic AI}
\affil[2]{SAFER}
\affil[3]{Trafikverket}
\affil[4]{Volvo Cars}
\affil[5]{Department of Computer Science and Engineering, Chalmers University of Technology}
\affil[1]{\textit {\{yinan.yu, samuel.scheidegger, jorg.bakker\}@asymptotic.ai}}
\affil[2]{\textit {\{johnfredrik.gronvall, erik.svanberg\}@chalmers.se}}
\affil[3]{\textit {magnus.palm@trafikverket.se}}
\affil[4]{\textit {johan.amoruso.wennerby@volvocars.com}}
\affil[5]{\textit {yinan@chalmers.se}}
\date{September, 2021}
\maketitle

\begin{figure*}[h!]
    \centering
    \includegraphics[width=0.8\textwidth]{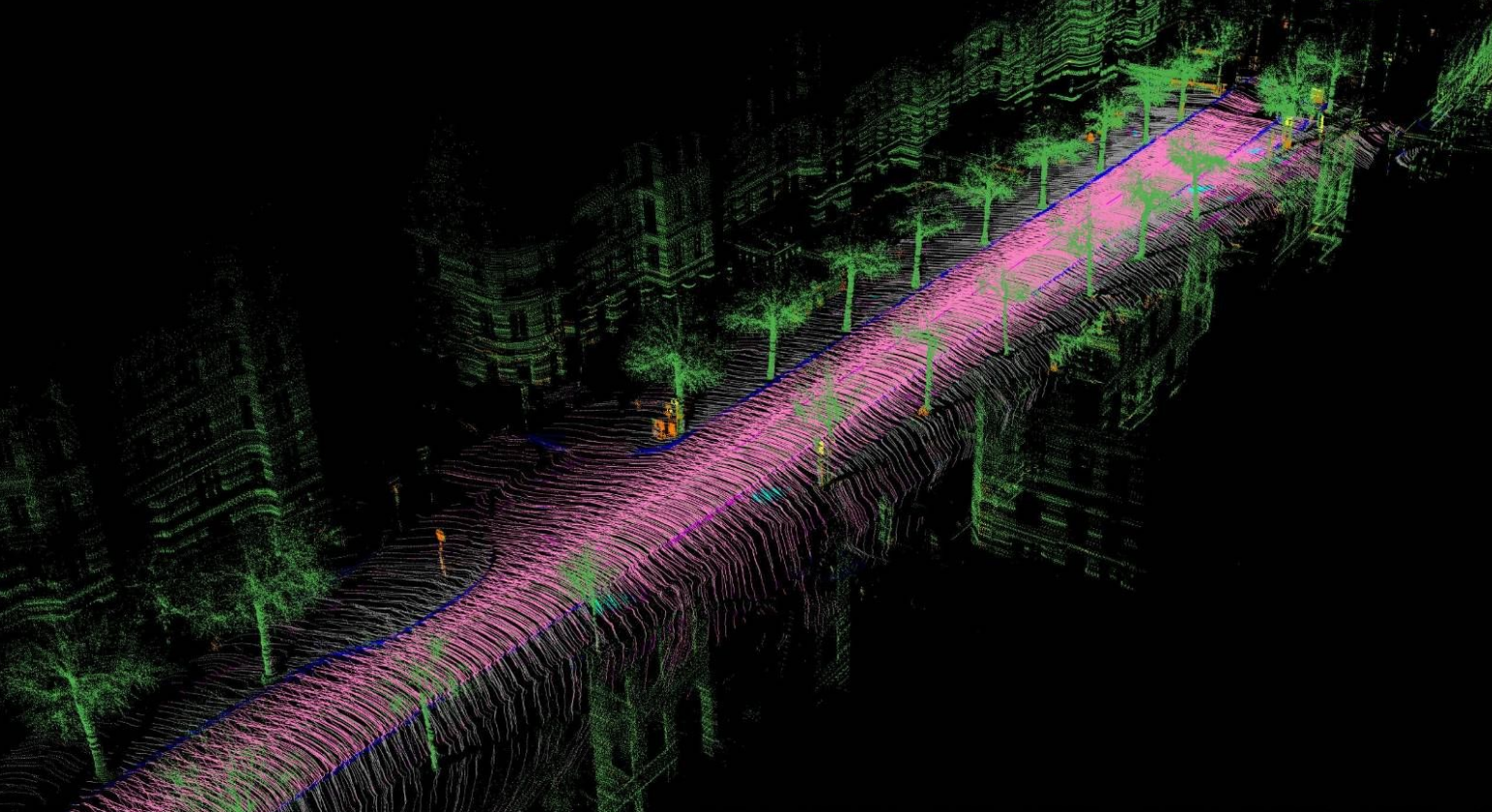}
    \alt{A sample outcome of roadside object detection on the data collected using the REVERE vehicle. }
\end{figure*}
\clearpage

\begin{abstract}

  Single-vehicle accidents are the most common type of fatal accidents in Sweden, where a car drives off the road and runs into hazardous roadside objects. Proper installation and maintenance of protective objects, such as crash cushions and guard rails, may reduce the chance and severity of such accidents. Moreover, efficient detection and management of hazardous roadside objects also plays an important role in improving road safety.
  To better understand the state-of-the-art and system requirements, in this pre-study, we investigate the feasibility, implementation, limitations and scaling up of data processing pipelines for roadside object detection. In particular, we divide our investigation into three parts: the target of interest, the sensors of choice and the algorithm design.
  The data sources we consider in this study cover two common setups: 1) road surveying fleet - annual scans conducted by Trafikverket, the Swedish Transport Administration, and 2) consumer vehicle - data collected using a research vehicle from the laboratory of Resource for vehicle research at Chalmers (REVERE).
The goal of this report is to investigate how to implement a scalable roadside object detection system towards safe road infrastructure and Sweden's Vision Zero.

\end{abstract}
 \vspace{5cm}

\clearpage
\section{Introduction}
According to the WHO Global Burden of Disease (GBD) model, there will be 2.4 million road deaths per year by 2030 unless immediate measures are taken. There are three main contributing factors to road safety: road users, vehicles and road infrastructure, among which road design, maintenance and roadside object awareness plays an important role.

Single-vehicle accidents are the most common type of accidents that leads to fatal outcomes in Sweden. What usually happens is that a car for some reason drives off the road to the left or right and overturns or runs into some inflexible object.

The infrastructure measures used to improve the road side area are mainly removing dangerous objects or setting up guard rails. Center railings are also important because almost half of the exits take place on the left.

To achieve well designed and maintained road infrastructure, manual road inspection is the traditional approach. For instance, the iRAP survey manual \cite{irapsurveymanual} describes a standard  data collection procedure for road inspection in order to star rate road segments in terms of how safe they are. However, for the purpose of large-scale road inspection on a regular basis, manual assessment is too time consuming, costly and potentially resulting in inconsistent outcomes. With the rapid development of perception and data processing systems, automated inspection approaches became available over the last two decades and started to gradually replace manual processes.

In this document, we summarize some of the existing approaches with respect to 1) the target of interest, 2) the sensors used for data collection and 3) the algorithms for data analysis. We then briefly describe the state-of-the-art solutions within each category and their relevance to our use case. The limitations of each approach are also discussed as a part of the document to show potential risks. Most of the computing techniques mentioned in this report are in the public domain in the form of research resources or open source softwares. We also briefly summarize some available commercial solutions that are relevant to our interests. These interests are mainly based on the road specifications from Trafikverket, the Swedish Transport Administration \footnote{The Swedish Transport Administration (Trafikverket): \url{https://www.trafikverket.se/}}, and the road inspection manual from the International Road Assessment Programme (iRAP) \footnote{iRAP: https://irap.org/}.

Road scanning data is one source that could provide an opportunity to analyze and describe the roadside status and attributes. The main data sources mentioned in this report are 1) the annual scan of paved state road network conducted by Trafikverket and 2) data collected \footnote{The data is collected using the research vehicle provided by the Resource for vehicle research at Chalmers (REVERE) lab (\url{https://www.chalmers.se/sv/forskningsinfrastruktur/revere/Sidor/default.aspx}). The collection and annotation are implemented by Asymptotic AI (\url{https://www.asymptotic.ai}).} on the public road in the Gothenburg area in Sweden.

The objective of this document is to investigate how to implement a scalable data processing pipeline to detect roadside objects towards safer road infrastructure and contribute to Sweden's Vision Zero \cite{visionzero}.

\section{Overview}

Automated and semi-automated road inspection can be achieved with a road infrastructure management system, which is a system consisting of a set of tools that automate or assist human actors to identify cost effective road maintenance and design strategies. There are two main steps involved: detection and assessment. {\bf Detection} refers to the process of perceiving and analyzing the condition of the road using various types of sensors and data processing techniques, whereas {\bf assessment} concerns the decision making process based on the discrepancy between the road specification and the observation. In this report, we mainly focus on the detection system. The assessment is considered the next step given the detection results.

\subsection{Detection}

There are three main building blocks in the detection system: the {\bf target} of interest, the choice of {\bf sensors} and the {\bf algorithms} for data analysis and processing.

\subsubsection{Target}

\begin{figure}[h!]
    \centering
    \begin{subfigure}[b]{1\textwidth}
        \includegraphics[width=\textwidth]{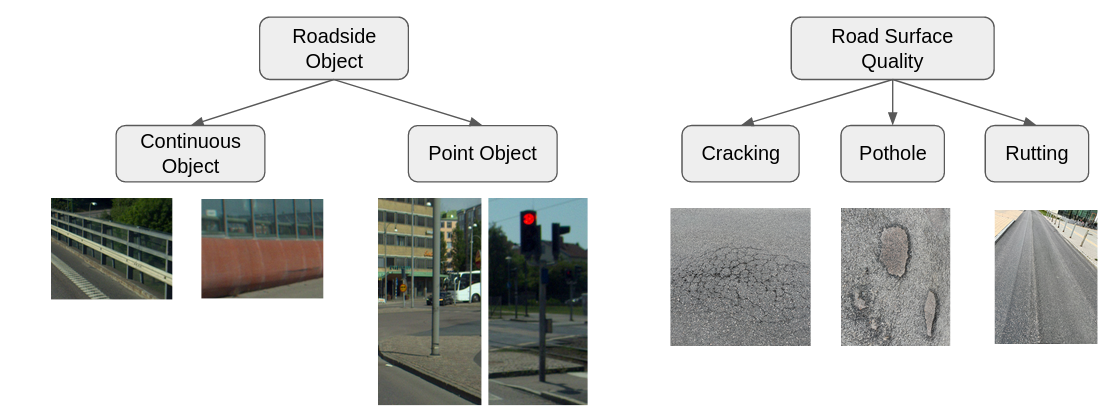}
    \end{subfigure}
    \caption{The target of interest for the detection task can be categorized into road surface condition and roadside objects.}
    \alt{Left: Roadside object detection is divided into three sub-categories: 1) debris, 2) continuous object (with example images of guard rails, rigid barriers, bridge pillars), and 3) point object (with example images of a traffic light and a pole). Right: Road surface quality with example images of cracking, pothole, rutting and shoving.}
    \label{fig:applications}
\end{figure}

The target of interest refers to what needs to be detected. There are two main categories in this context: the road surface condition and roadside objects detection. Some examples can be found in Fig.~\ref{fig:applications}.

\paragraph{Road surface}

Maintaining a good road surface condition is a crucial step towards high quality road infrastructure. Typically, the surface condition includes but is not limited to the following aspects:
\begin{itemize}
\item Pavement distress \cite{coenen2017review,ragnoli2018pavement, schnebele2015review}: e.g. pothole, cracks, rutting, rumble strips
\item Marking quality: e.g. unclear lane marking
\item Surface friction: e.g. icy or slippery road
\item Serviceability/smoothness: deviations in pavement surface

\end{itemize}

\paragraph{Roadside objects}

According to iRAP, there are eight common types of crashes, where the run off road crashing type is highly correlated with the presence of hazardous roadside objects or missing protective infrastructure. These non-compliant roadside objects can significantly increase fatal injury. Some examples of non-compliant roadside objects include:
\begin{itemize}
\item Broken road sign
\item Missing crash barrier
\item Unexpected obstacle on the road
\item Animals on the road
\item Misplaced objects that may block the driver’s vision
\item Poorly managed construction site, e.g. oily road, poor traffic control, misplacement of objects, etc.
\end{itemize}

In this report, we mainly focus on the task of roadside object detection.

\subsubsection{Sensors}
\label{sec:sensors}
Given the objective of automatically detecting roadside objects, the sensors of choice for data collection play a key role in the overall performance of the system.
Particularly, in this context, a typical set of features to be extracted include
1) the structure of the target object in the three dimensional space,
2) semantics of the measurement and
3) geographical information of the target.
The sensors need to be able to measure relevant information such that these features are captured.
In this section, we introduce some commonly used sensors for this type of applications.

\paragraph{GNSS receiver}
\begin{itemize}
  \item{{\bf Description:}} GNSS (GPS, Galileo, BDS, etc) sensors provide critical information about the global position and timestamp for the data collection system. Depending on the accuracy, this information can be used for various purposes, including localization, time synchronization, reverse geocoding, route planning, etc. Moreover, it is a common practice for other sensors to be equipped with a built-in GNSS device for georeferencing and data correction. This sensor is typically considered a necessity to a road inspection system.
  \item{\bf Risk factors:}
    \begin{itemize}
    \item Reliability: In certain areas, such as inside a tunnel, the GNSS receiver may lose contact with the satellites and result in poor signal reception.
    \item Robustness: The GNSS receiver device often has built-in post-processing algorithms to estimate the exact location of the ego carrier. These algorithms may have edge cases on certain measurements. This factor needs to be taken into consideration during the data processing step.
    \end{itemize}
  \end{itemize}
  \paragraph{Ego motion sensor (gyroscope/speedometer/accelerometer)}
  \begin{itemize}
  \item{\bf Description:} In a typical setup, the data collection carrier is moving during the collection process and hence its ego motion needs to be compensated in order to extract information from the measurement as accurately as possible. To achieve this, ego motion sensors are typically installed in the data collection system. These sensors refer to, for instance, a gyroscope to measure the heading, pitch and roll of the ego carrier, a speedometer to estimate the speed, an accelerometer to provide the acceleration information, etc. Some of these sensors are often combined into one single Inertial Measurement Unit (IMU) unit to provide more accurate estimates.
  \item{\bf Risk factor:} The raw measurements generated by an IMU platform may not be reliable due to the potentially high noise level resulting from, for instance, vibrations and time drifts. This is especially problematic for sensors with low manufacturing cost.
    To reduce the impact of such noise, filtering techniques are often applied. Depending on the manufacturer, this postprocessing step may be included as part of the hardware configuration or it may be a separate data processing step after the sensor generates the output. If a high resolution is required by the use case, users shall pay close attention to the data collection and processing pipeline of the platform and watch out for potential artifacts generated by the filtering steps.
  \end{itemize}
 \paragraph{Camera (2D)}
 \begin{itemize}
 \item{\bf Description:} Images are commonly used for object detection and classification. 2D camera images are the most adopted among different types of cameras. The reason for using 2D cameras is often associated with their relatively low cost, high availability and flexibility. In particular, 2D cameras are frequently used for detecting pavement distress, such as potholes and cracks \cite{yang2019feature, maeda2018road, zhang2016road}, since the two dimensional characteristics are often sufficient for the detection task. Furthermore, over the last two decades, thanks to the rapid development of deep learning techniques, object detection and classification have shown increasing accuracies and performances using 2D images, which has further increased the popularity of 2D cameras in automated systems.
 \item{\bf Risk factors:}
   \begin{itemize}
   \item Environmental factors: Lighting and weather condition can have a significant impact on the quality of the raw data.
   \item Hardware characteristics: Digital artifacts, such as rolling shutter may cause wobble, skew or aliasing. Low sensitivity sensors or low light lenses might cause motion blur in low light conditions. Images from low quality lenses with heavy distortion might be hard to correct.
   \end{itemize}
 \end{itemize}
\paragraph{LiDAR}
\begin{itemize}
\item{\bf Description:}
Light Detection And Ranging (LiDAR) \cite{dong2017lidar} is a remote sensing technology for measuring the distance of objects using a light transceiver. There are mainly two types of LiDAR installation setups: airborne and terrestrial. Airborne LiDAR is mounted on a flying aircraft, such as a helicopter or a drone, to create a 3D model of the landscape, whereas terrestrial LiDAR is installed on the ground. Furthermore, terrestrial LiDAR can be stationary or mobile. For road safety applications, mobile LiDAR is often used.

In particular, LiDAR sensors for the passenger vehicle industry have so far mostly been used by map scanning vehicles and by development vehicles for autonomous driving applications. Over the coming five years LiDAR sensors will be introduced on production vehicles and the availability of data from LiDAR sensors will therefore be dramatically increased over the coming years. Significant development is being performed to industrialise and improve LiDAR technology for vehicles and this is expected to lead to dramatically improved performance and deployment within production vehicles. Existing data sets from test vehicles could possibly be made available to a research project by OEMs, companies developing autonomous driving applications and map development companies.

The exact working mechanism of the LiDAR technology is beyond the scope of this report. Interested readers may refer to \cite{soilan2019review, dong2017lidar} for further elaboration on the subject.

In this report, we present some of the key aspects that may have a direct impact on our use case. Particularly, we focus on mobile terrestrial LiDAR systems due to their wide adoption in the automotive and transportation industry for perception systems on the road \cite{che2019object, WANG2020175, soilan2019review}.

\item{\bf Risk factors:}
   \begin{itemize}
   \item Environmental factors: Weather condition can have a large impact on the LiDAR output. To illustrate this, two examples are shown in Fig.~\ref{fig:lidar_weather},
     where Fig.~\ref{fig:snow} 
     shows a LiDAR sample collected in the snow and Fig.~\ref{fig:fog} in a foggy weather condition.
In Fig.~\ref{fig:snow}, the reflected laser beams on the falling snow flakes are clearly visible, which results in outliers and potential errors in the analysis. In Fig.~\ref{fig:fog}, on the other hand, the LiDAR measurements are barely visible except for some of the retroreflective lane markings that are close to the ego vehicle.
\begin{figure}[h!]
    \centering
    \begin{subfigure}[b]{0.53\textwidth}
        \includegraphics[width=\textwidth]{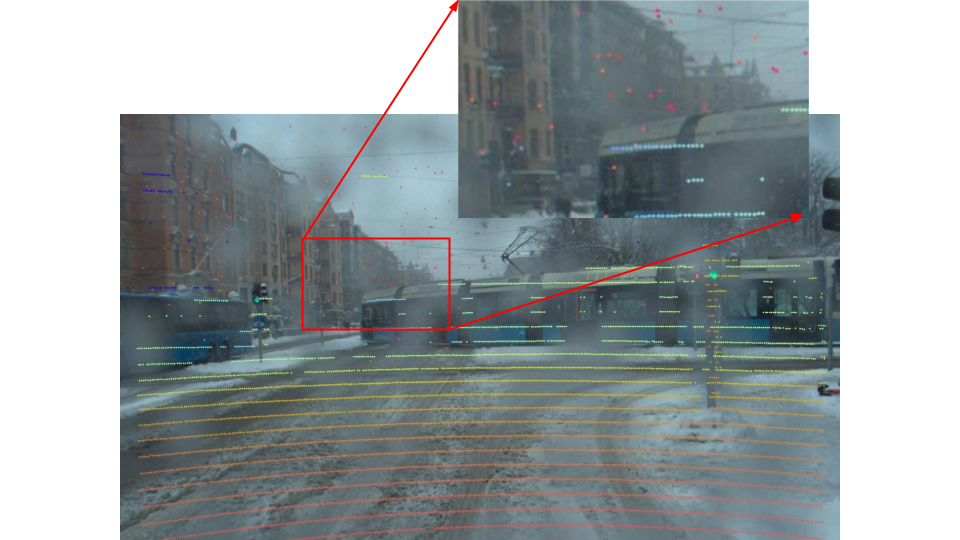}
        \caption{A sample LiDAR scan in the snow, where the snow flakes are clearly visible as part of the reflections.}
        \alt{Sample LiDAR scan in the snow, where the snowflakes are clearly visible.}
        \label{fig:snow}
    \end{subfigure}
    ~
    \begin{subfigure}[b]{0.43\textwidth}
        \includegraphics[width=\textwidth]{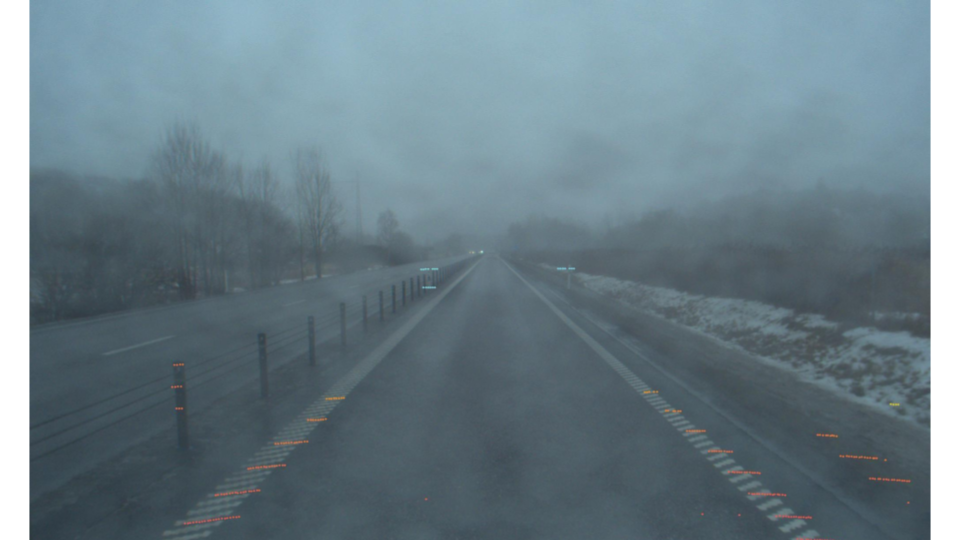}
        \caption{A sample LiDAR scan in a foggy environment, where only very few points are visible.}
        \alt{LiDAR measurements are barely visible in a foggy environment.}
        \label{fig:fog}
    \end{subfigure}
    \caption{Examples of how weather condition may have an impact on the LiDAR measurements. The LiDAR points are projected onto the corresponding camera image for visualization purposes.}\label{fig:lidar_weather}
  \end{figure}

   \item Limited range:
     The dynamic range of a reasonably priced LiDAR system can be limited. An example can be found in Fig.~\ref{fig:lidar_range}, where a red car is obviously visible in the camera image, but not in the LiDAR data.
     \begin{figure}[h!]
    \centering
    \begin{subfigure}[b]{0.6\textwidth}
        \includegraphics[width=\textwidth]{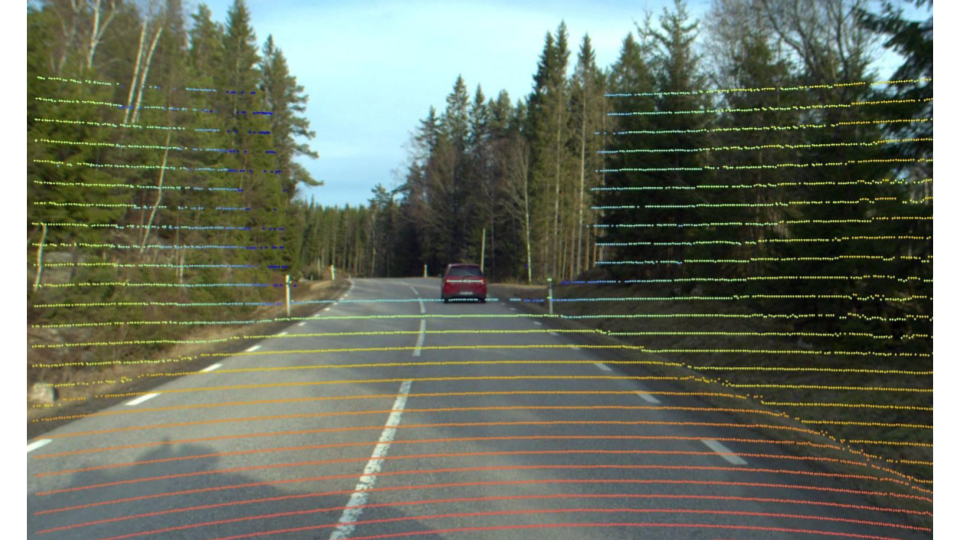}
    \end{subfigure}
    \caption{Illustration of the range limitation. In this example, a red car is in front of the ego vehicle and it is outside the visible range of the LiDAR sensor equipped on the ego vehicle. If this target vehicle keeps a constant or increasing distance with respect to the ego vehicle, it will never be visible by the LiDAR sensor.}
    \alt{Illustration of the range limitation. In this example, a red car is in front of the ego vehicle and it is outside the visible range of the LiDAR sensor equipped on the ego vehicle. If this target vehicle keeps a constant or increasing distance with respect to the ego vehicle, it will never be visible by the LiDAR sensor.}
    \label{fig:lidar_range}
\end{figure}
   \item High cost: most LiDAR sensors are considered high-end devices that have a relatively high cost.
   \item Potentially complex data processing pipeline: LiDAR scans contain rich three dimensional information of the surroundings. However, the data processing steps are typically more involved compared to, for example, data processing for images. This complexity can bring challenges to the detection system as a whole.
   \end{itemize}
 \end{itemize}

 \paragraph{System integration}
 \begin{itemize}
 \item{\bf Description:}
 \begin{itemize}
 \item{\bf Data logging:}
    The aforementioned sensor data needs to be received and recorded on a logging device during a data collection activity, which means that a complete logging solution needs to be in place.
  \item{\bf Data processing requirement:} Once the sensor data is received, the data processing and analysis for the detection can be either real-time or offline. Real-time processing requires a real-time system operating in the data collection carrier so that data streams can be processed as soon as they have arrived. Typically, for real-time systems, there are specifications on the reaction time. For our use case, however, we mainly focus on offline detection, which means that the requirement on the duration of data processing is much looser compared to real-time systems and we can utilize central computational resources to process the sensor data in order to achieve a good accuracy.
  \item{\bf Data fusion:} Data from multiple sensors need to be aggregated for detecting the target of interest. Generally speaking, there are three main challenges: time synchronization, sensor calibration and data association. Time synchronization refers to synchronizing the timestamps among different sensors. Sensor calibration is the process of corresponding measurements from one sensor to another. Data association typically refers to the step where detected objects from different sensors are being associated to one another.
  \end{itemize}
 \item{\bf Risk factors:}
   \begin{itemize}
   \item Time synchronization: It depends on the accuracy of the built-in GNSS device, the frame rate of each sensor, and the quality of the time synchronization mechanism from the logging system. If the time synchronization is not satisfactory given the system requirements, time adjustments need to be applied to compensate the time difference between sensors.
   \item Sensor calibration: There are two types of sensor calibrations: intrinsic and extrinsic. Intrinsic calibration refers to transformations that do not depend on the outside world, whereas extrinsic calibration concerns the mapping with respect to the world or between different sensors. Both procedures need to be carried out in order to properly interpret the sensor outcome. Depending on the quality of the sensor and time synchronization, the calibration may result in errors, which need to be carefully examined.
   \item Data association: The challenge of the data association step is mainly on the algorithmic level. The performance is usually bounded by the quality of the data and the algorithm of choice.
   \item Development data set: a biased data set for development inevitably causes bias in the detection system. In order to handle and evaluate the system for a wide range of situations, ensuring a large data variety is one of the key measures to take. To this end, the development data set shall consist of data collected in different scenarios, for example, time of day, various weather conditions and seasonal variations.
   \end{itemize}
\end{itemize}

  \paragraph{Other sensors} There are other types of sensors being used for object detection systems. For example, 3D/stereo cameras, line scan cameras, radars, acoustic sensors, etc. These sensors can be used to provide additional information. However, the necessity of introducing more sensors needs to be analyzed to avoid too high system complexity. 

  \subsubsection{Algorithms}

  \paragraph{Deep learning based object detection}
  \begin{itemize}
  \item{\bf Description} Deep learning algorithms are widely used in robotics and automotive perception systems. Within this context, one of most successful applications are object detection, classification and semantic segmentation in images.
The working mechanism and implementation of deep learning is outside the scope of this document. What we focus on is the potential choice of deep learning algorithms based on their strength and weaknesses for roadside object detection and how to combine these algorithms with more traditional model-based approaches. This topic is elaborated throughout the document.
  \item{\bf Risk factors} Despite their popularity, there are several risk factors associated with deep learning.
    \begin{itemize}
    \item Black box: There are increasing efforts on interpreting and understanding the outcome of deep learning algorithms, but this area of research is still very limited. With the tooling available today, such as transfer learning and advanced programming libraries, it is straightforward to obtain some automated results as a starting point. However, the black box nature of deep learning brings challenges when adjustments are needed in order to improve the performance. This is an open research question that needs to be addressed case by case.
    \item Hyperparameter tuning: In order to improve the performance and robustness of the algorithms, extensive hyperparameter tuning needs to be performed as a combinatorial optimization problem. For a typical object detection system, there are many hyperparameters involved, which makes the task challenging. This process can be carried out either manually or (semi-)automatically. Either way, the outcome of the hyperparameter tuning needs to be stored and sorted for interpretation, optimization and reproducibility.
      Overall, with the state-of-the-art techniques, it is still very challenging to breakthrough a performance bottleneck of a neural network.
    \item High costs: Deep learning requires specialized hardware infrastructure such as GPUs and fast storage. Both the initial and running costs (e.g. IT support, server hosting, cooling, electric power, etc) are considerably higher than traditional data processing pipelines. When the data set is small enough, multiple services can be found available for free for a limited training time. However, these supports are not sufficient for building a production ready roadside object detection system, where investments on the infrastructure is inevitable. Due to its high cost, this investment needs to the strategized.
    \item Slow feedback and long development time: The feedback loop for deep learning algorithm development is usually slow due to the long processing time of generating annotations, hyperparameter tuning and potentially large amounts of data to traverse during training. Efficient hardware and software infrastructure may be used to help speed up the development process.
    \item Requiring a large amount of training and validation data: Data needs to be collected and annotated for training and validation. Historically, this process has been largely depending on expensive manual processes. Recent developments have shown promising results with automated annotation processes to replace human labors.
    \end{itemize}
    One way to mitigate these risks is to combine deep learning algorithms with other more established model-based techniques.
    Utilizing smart hardware and software infrastructure is another key aspect to achieve high efficiency.
  \end{itemize}
  \paragraph{Image processing}
  \begin{itemize}
  \item{\bf Description:} Image processing techniques have been widely used for various applications such as camera calibration, feature extraction, pattern recognition and classification. Most of the traditional image processing techniques are linear methods, which reply on operations such as projections, convolutions and other more generic linear transformations and decompositions. While these well studied techniques are efficient for certain use cases, they are usually outperformed by deep learning when analyzing complex, localized semantics in images such as object detection. However, due to their simplicity and maturity, it is still preferred to use image processing algorithms whenever applicable.
  \item{\bf Risk factors:} Image processing techniques have shown high performance when it comes to structural analysis of the imagery. However, they have limited capabilities for complex semantic analysis. For example, the contour of a crack on the pavement can be accurately detected, but the algorithm may fail at understanding the semantic of the area as a crack. A second risk is that the effectiveness of most image processing techniques heavily rely on the quality of the image, such as the resolution, contrast, Signal-to-Noise-Ratio (SNR), etc, which makes them not robust enough for many applications and use cases.

    The choice of the camera sensor has a large impact on the algorithm outcome. Unfortunately, the budget may not allow purchasing cameras with satisfactory qualities. In this case, one shall be mindful of the limitations from the hardware.

   In practice, image processing algorithms can be combined with other techniques, where image processing is typically used as a pre-processing step to utilize the domain knowledge of the sensor. Machine learning techniques, in particular, deep convolutional neural networks, are a popular choice to analyze the semantics of images.

  \end{itemize}

  \paragraph{Model-based algorithms}
  \begin{itemize}
  \item{\bf Description}
    This type of data processing algorithms refer to techniques with a relatively strong assumption on the underlying model of the data generation. The subsequent analysis is largely based on this underlying assumption. The advantage of such techniques is that it is typically possible to analyze the asymptotic behavior of the algorithm, such as it consistency, upper/lower bounds of the accuracy, etc. Their performances are often easier to interpret due to the explicit model assumption.
    \item{\bf Risk factors}
      The strong assumption on the model can be inaccurate, which may lead to catastrophic failure in the detection outcome.

      To improve the model selection, it is advised to first conduct thorough exploratory analysis on the data. Moreover, the coverage of data representation plays an important role and therefore one needs to collect data in different circumstances to cover as many scenarios as possible. It is also advised to use multiple models to reduce the bias in general.
    \end{itemize}

When it comes to algorithm design, one shall be aware of the bias-variance tradeoff. Loosely speaking, bias can be understood as how accurate the algorithm is on average, whereas variance can be considered as how stable the algorithm is. Algorithms with low complexity tend to provide a highly biased solution with relatively low variance and vice versa. For example, deep learning algorithms with a large amount of trainable parameters are considered highly complex, which may result in an outcome with a large variance. This effect can be reduced by increasing the size and coverage of the training data set. Moreover, when combined wisely, these algorithms can compensate for the pitfalls from one another and produce a more stable outcome with a satisfactory high accuracy.


\subsection{Assessment}
Based on the detection results, decisions need to be made depending on the use cases. These use cases can be, for instance, predictive maintenance, road design, route planning, etc.
These applications involve a decision making step after the detection is received and interpreted. In this document, we skip the assessment process since this subject is not our main focus. However, the goal and requirements of the assessment step shall be taken into consideration when developing the detection system.

\section{Roadside Object Detection Pipeline}
Poorly maintained roadside objects pose a fatal threat to the safety of the drivers. In the previous sections, we have introduced the building blocks for constructing a roadside object detection system with the objective of improving road infrastructure.

In this section, we put together all these building blocks and describe the system as a whole. We mainly focus on the detection of roadside objects using LiDAR, GNSS and 2D camera sensors with deep learning, image processing and model-based methods.

\subsection{Methodology}
\begin{figure}[h!]
    \centering
    \begin{subfigure}[b]{0.7\textwidth}
        \includegraphics[width=\textwidth]{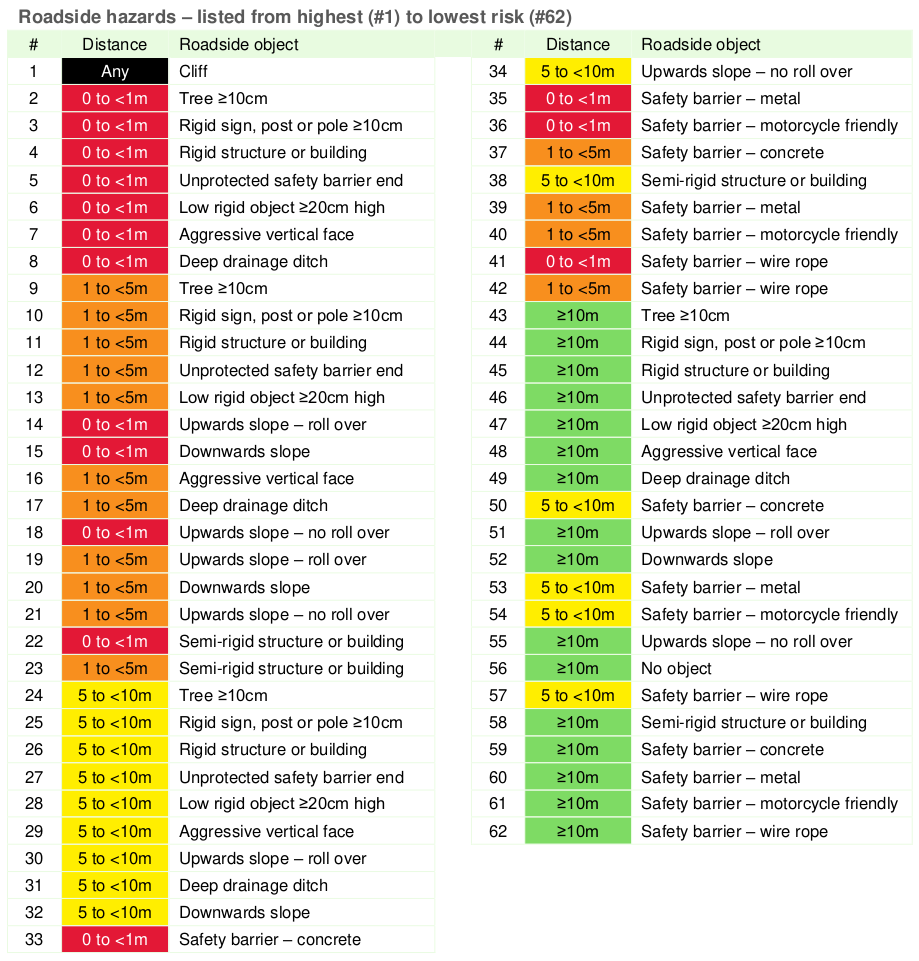}
    \end{subfigure}
    \caption{A ranked list of hazardous roadside objects from iRAP.}
    \alt{A ranked list of hazardous roadside objects from iRAP. The top item is cliff, followed by trees, rigid sign, post or pole that are closer than 1 meter and larger than 10 cm in diameter. According to this list, the least dangerous items are safety barriers.}
        \label{fig:roadside_irap}
\end{figure}

\paragraph{Target of interest: what needs to be detected?} According to iRAP \cite{irapsurveymanual}, there is a list of hazardous roadside object types shown in Fig.~\ref{fig:roadside_irap}. These object types are ranked based on how risky they are from a road safety perspective.

Another important source is a list of roadside objects from Trafikverket, which are classified into three categories:
\begin{itemize}
\item Side area geometry
  \begin{itemize}
  \item Incline of the inner slope
  \item Slope of the outer slope
  \item Distance to the bottom of the ditch
  \item Height difference between road edge and ditch bottom / surrounding ground
  \item Slope at intersection / connection road
  \end{itemize}
\item Dangerous object in the side area
  \begin{itemize}
  \item Type of dangerous object (tree, mountain, pole, deep water, precipice, etc.)
  \item Distance to dangerous objects
  \end{itemize}
\item Protective devices
  \begin{itemize}
  \item Side railing (railing type, railing end, etc.)
  \item Center railing (railing type, railing end, etc.)
  \item Crash cushion
  \end{itemize}
\end{itemize}
From a methodology point of view, these targets of interest can be divided into {\bf point objects} and {\bf continuous objects}. Point objects refer to poles, trees and posts, whereas continuous objects include barriers, guard rails, rigid walls, etc.
The annotations and detection techniques for these two types of objects may differ due to their structural differences.

\paragraph{Sensors: 2D or 3D?} Certain types of pavement deterioration, such as potholes and cracking, can be identified from 2D camera images due to their distinct characteristics compared to the ambience. However, the detection and classification of roadside objects usually relies on 3D information, since their features are better represented in the three dimensional space \cite{WANG2020175}. The most commonly used 3D sensor for roadside object detection is the LiDAR sensor. Camera images are often combined with the LiDAR sensor for classifying the objects given their visual characteristics \cite{balado2020transfer}.

\paragraph{Algorithms: model-based or deep learning?} There are two mainstream categories of algorithms: model-based methods and deep learning algorithms. The main difference is that model-based algorithms impose strong assumptions in the data generation process, whereas the construction of deep learning models is data-driven, meaning that it requires little to none prior knowledge about the data from the human actors; the model is largely determined by the available training data. Generally speaking, model-based algorithms have a tendency of resulting in a relatively large bias when the assumptions are inaccurate, whereas deep learning algorithms tend to have a large variance in their outcomes when the training data set is small. Compared to model-based techniques, deep learning is less analytical in general due to the lack of theoretical assumptions. However, deep learning often shows high exploratory capacity given a large amount of training data and detects patterns that are not understood by humans.
Due to this characteristic, deep learning methods are widely used for complex data analysis tasks, such as object detection and classification in images.

\paragraph{Deep learning: prerequisites?} There are two main prerequisites when it comes to developing deep learning algorithms: annotated data and computational power.

Deep learning is data-driven, meaning that data is at the core of the algorithm development. In this context, the concept “data” includes the information of the sensor measurement itself and the expected outcome of the detection system. For instance, it may contain a picture of a bridge pillar and some meta information indicating what it is - a bridge pillar - and where it is in the image. This meta information is called annotation in deep learning development and it is used for training the model to recognize and locate bridge pillars in images. Traditionally, annotations are made by human annotators. For instance, they may draw a box around the object and add a tag “bridge pillar” to indicate what is contained in that box. However, this process is labor intensive, which is typically time consuming, hard to scale and economically inefficient. Over the last five years, automated annotation has become a popular alternative to its manual counterpart \cite{yang2021auto4d, ettinger2021large}. In the context of automotive and road safety applications, the automation is typically based on sensor fusion, especifically the combination of the geographical information, the 3D scans from the LiDAR and 2D camera images. By combining these sensors, the annotation process can be automated to a large extent and hence the cost of annotation can be significantly reduced.

A second dependency of deep learning algorithm development is the computational power, for instance, the GPU compute infrastructure. To be able to develop in-house deep learning algorithms, the investment of either renting or buying GPU infrastructure is inevitable. The budget of the initial cost and the running cost of deep learning infrastructure needs to be estimated before a project starts and iterated as the project develops.

As of today, there are mainly three ways to utilize deep learning infrastructure:
1) self-hosting,
2) cloud-based services and
3) collocation hosting.
Generally speaking, self-hosting has the lowest cost for the hardware but the running cost can be significantly higher due to the responsibility of taking care of the physical space for the servers, hardware maintenance and software development.
Cloud-based services are highly accessible with the lowest initial costs, but the running cost for customer service and data transfer can be very high when scaling up. Moreover, servers are owned by the providers, which may cause security and compliance issues. Another potential bottleneck is that the bandwidth might limit the efficiency of data transfer if the network connection from the user to the data center is not satisfactory.
Collocation hosting services can cost more in the beginning compared to self-hosting, but they typically result in a much lower running cost. Moreover, the provider does not own the servers, which means that the users have better control over how data and servers are managed. However, the servers do not necessarily share the same location as the end user, which may cause issues if fast network connections cannot be drawn. A mixture of all three approaches is also possible depending on the flexibility of the provider.

The infrastructure set up is a significant part of the data processing pipeline.
The decision shall be made based on the budget, the available in-house expertise and the specifications from the developers and use cases.

\paragraph{Expected algorithm output: what should they be?} Before choosing an algorithm, one needs to define the expected outcome of the detection system and decide how they should be annotated. For each type of sensors, there are commonly used types of annotations:
\begin{itemize}
\item Distributed objects: semantic segmentation, panoptic segmentation
\item Point objects: 2D/3D bounding boxes, polygon, panoptic segmentation
\end{itemize}
These annotations represent the expected output of the algorithms, but they do not necessarily reflect the requirement of the end use case. For instance, a ditch that is more than one meter in depth and one meter in diameter is expected to be detected. Assume that the algorithm has successfully located a ditch in the LiDAR point cloud, a post processing step is still needed afterwards to extract the exact dimension of the ditch, which is not usually included in the standard annotations. This post processing step is typically not data-driven, but it is the downstream analysis to the annotation and the algorithm output. Hence, it needs to be taken into consideration when designing the specifications of the annotations to make sure that necessary information is reflected in the annotation.

\paragraph{Evaluation: what is considered a good system?}
To better understand the requirements of the end product and effectively analyze the system as a whole, we propose a set of evaluation criteria for the detection system:
\begin{itemize}
\item Performance: The performance refers to the quality of the system output, e.g. precision/recall of the roadside object detection system.
\item Specifications: The specifications of the end use case needs to be satisfied. For instance, if the specifications require the system to detect a ditch that is one meter deep, a failure case would be, for example, when the algorithm detects a ditch but failed to understand the depth of the ditch.
\item Processing speed and scalability: The algorithms need to be developed and validated within a reasonable time frame. The inference time of the developed system needs to be taken into account when designing the system.
\item Resources costs: The cost for resources is limited by the budget for the project. The cost includes initial costs and running costs for upgrades and maintenance. The costs for resources include but are not limited to:
  \begin{itemize}
  \item Storage servers: slow storage for raw data, fast storage for deep learning training and data caching
  \item Compute servers: CPU servers for preprocessing and GPU servers for deep learning
  \item Hardware infrastructure hosting, cooling, energy consumption
  \item Human resources: IT services, algorithm development, data annotation, human inspection to provide domain knowledge
  \end{itemize}
\end{itemize}

\subsection{State-of-the-art}
\label{sec:sot}
Roadside object detection can be roughly divided into image-based systems and 3D LiDAR point cloud analysis.
In this section, we briefly summarize the state-of-the-art for roadside object detection by introducing commonly used annotations, models and their potential risks. We mainly focus on the risks that have a direct impact on the performance of the system as a whole. For more limitations and risks of the sensors themselves, please refer to Section \ref{sec:sensors}.

\subsubsection{Image-based data processing}
\paragraph{Annotation}  There are a number of commonly used annotations for object detection in images, among which, the most popular annotations are 2D bounding boxes, 3D bounding boxes, semantic segmentation and panoptic segmentation. The difference between semantic segmentation and panoptic segmentation is that the former only contains the object type of each pixel while the latter also groups together all pixels that belong to one object.

As mentioned before, the annotations can be acquired by a manual process or a (semi-)automated process. In the application of roadside object detection, the automation typically relies on the 3D information from a time synchronized LiDAR sensor.

\paragraph{Algorithms} Convolutional neural networks are the most popular choice due to their superior capability of extracting features from images. For each of the aforementioned annotations, there are multiple specialized open source algorithms designed for extracting the target information. For instance, YOLO, VGG, FasterRCNN, etc are popular choices for detecting 2D bounding boxes and Mask-RCNN is commonly used for segmentation of each object instance. Depending on the licensing, some of these models can be adopted off-the-shelf, while an additional training step is still required to achieve a desirable performance. A common practice is to use a pre-trained version of the model as a starting point and use transfer learning to continue training them on the specific in-house data sets. By using a pre-trained version in combination with transfer learning, the training time and cost can be significantly reduced compared to developing a deep learning model from scratch. For commercial users, it is necessary to careful examine the licensing model of the free software and pre-trained neural networks.

\paragraph{Limitations} One limitation of image-based detection is that it does not inherently contain 3D information of the environment and the objects of interest. While it is possible to train a neural network to estimate the 3D position and dimensions of an object, the estimation is usually imprecise. Moreover, due to the lack of 3D information, occlusion of objects is a frequently encountered challenge. This issue is usually compensated by combining other sensors, such as a laser scanner, together with the camera to provide sufficient information for the end use case. Another risk is the challenge of detecting small objects. Small objects are defined as objects that span very few pixels in an image and they may pose serious threats to the road users despite their insignificant appearance. This is an open research topic in the field of image processing and deep learning. For example, in \cite{CG20} the authors have proposed some techniques to enhance the accuracy of small object detection.

\subsubsection{LiDAR-based data processing}

The LiDAR sensor is often used as a source of reference due to its capability of building a highly accurate 3D model of the surroundings for roadside object detection \cite{ma2018mobile}. Because of its importance, we first give a relatively detailed description of the LiDAR sensor to better strategize the system design.
The type of LiDAR sensors we focus on is the mobile terrestrial system, which means that the LiDAR is mounted on a moving vehicle on the ground.

\begin{figure}[h!]
    \centering
    \begin{subfigure}[b]{0.485\textwidth}
        \includegraphics[width=\textwidth]{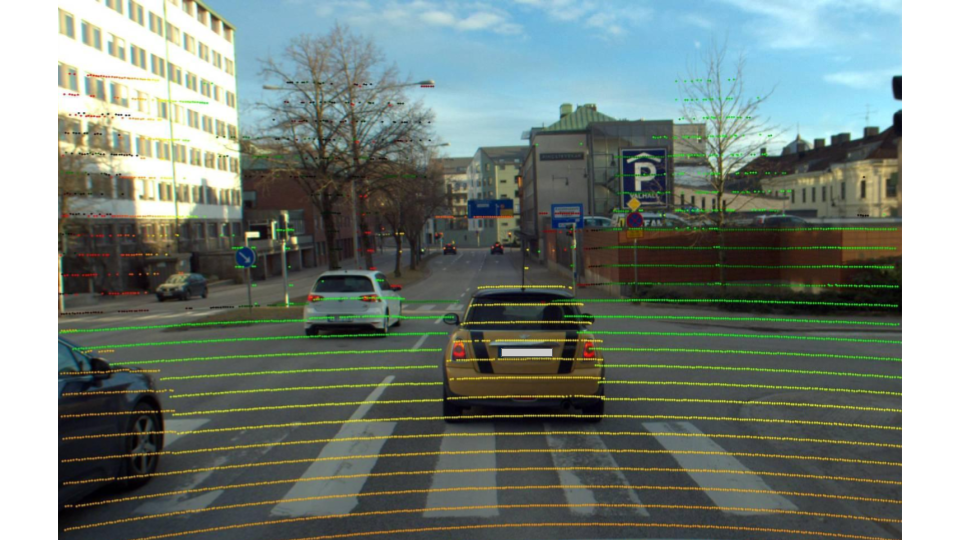}
        \caption{A LiDAR scan measured in a good weather condition. The points are projected onto the image recorded by a time synced forward looking camera.}
        \alt{Example of a LiDAR scan measured in a good weather condition. The LiDAR points are projected onto the image recorded by a time synced forward looking camera. It is shown that as opposed to a 2D camera image, the LiDAR is capable of providing distance information.}
        \label{fig:lidar_2d}
    \end{subfigure}
    ~ 
    \begin{subfigure}[b]{0.485\textwidth}
        \includegraphics[width=\textwidth]{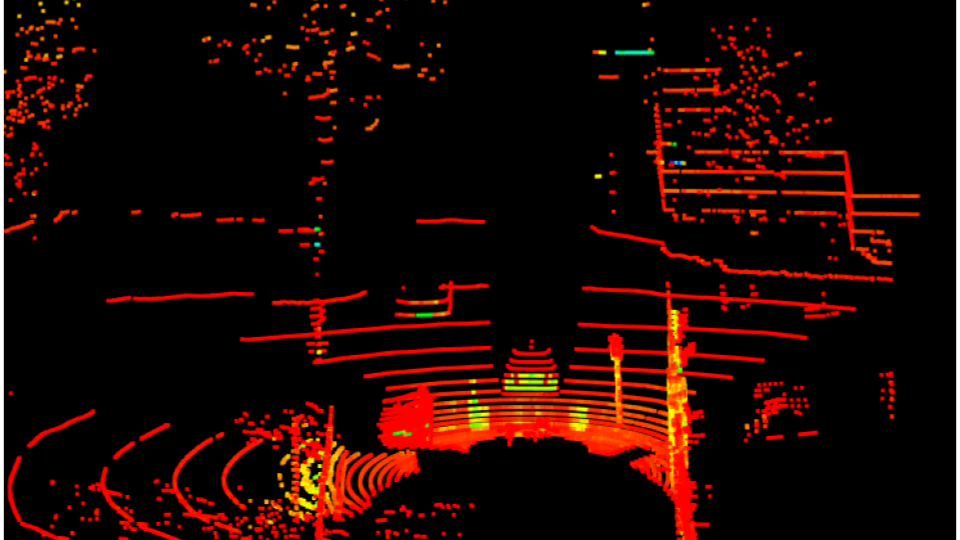}
        \caption{The same LiDAR measurements shown in the three dimensional space. The colors of the points indicate the intensity of the reflections.}
        \alt{LiDAR measurement in 3D. The color reflects the reflectivity.}
        \label{fig:lidar_3d}
    \end{subfigure}
    \caption{Example of a sample LiDAR scan.}\label{fig:lidar_output}
\end{figure}

\paragraph{Sensor output} A typical LiDAR sensor returns two measurements of a surface in the surroundings: 1. the 3D location (x, y, z) relative to the ego vehicle, and 2. the intensity of the reflected laser beam. For a well designed LiDAR system, the spatial resolution can be up to centimeter accuracy, which is sufficient for most applications. However, the theoretical resolution can be compromised by how the light beam is reflected. There are multiple factors that can influence this reflection, which in turn have a significant impact on the intensity produced by the LiDAR.

\paragraph{Influencing parameters of intensity} To better understand the processing steps of the LiDAR data, we briefly introduce the influencing factors of the intensity measurement and what can be done to circumvent potential pitfalls. In \cite{kashani2015review}, the authors have summarized four most important effective factors that influence the intensity: 1) target surface characteristics, 2) data acquisition geometry, 3) instrumental effects and 4) environmental effects. Instrumental effects are inherent features from the LiDAR system design. A comparison of ten different LiDAR models from multiple manufacturers (Velodyne, Hesai, Ouster, RoboSense) for the application of automated driving can be found in \cite{lambert2020performance} for interested readers. For a given LiDAR instrument, we mainly address the impact of roadside objects and road surfaces.

\begin{figure}[h!]
    \centering
    \begin{subfigure}[b]{0.8\textwidth}
        \includegraphics[width=\textwidth]{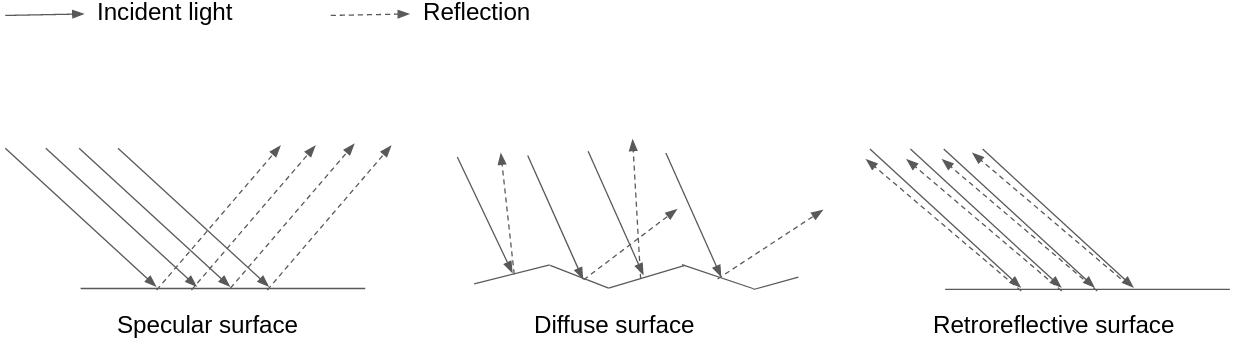}
    \end{subfigure}
    \caption{Illustration of how light is reflected on different types of surfaces.}
    \alt{Illustration of how light is reflected on specular surface (mirror like reflection), diffuse surface (reflected at many angles due to the roughness of the surface) and retroreflective surface (reflected back to its origin).}
    \label{fig:reflection}
\end{figure}

Particularly, we illustrate three most relevant types of reflective surfaces (see Fig.~\ref{fig:reflection}):

\begin{itemize}
\item{\bf Specular surface}
  \begin{itemize}
  \item Description: a surface with smooth, mirror-like reflectance
  \item Example: icy pavement, water, mirror, glass window
  \item Detection difficulty: hard
  \end{itemize}
\item{\bf Diffuse surface}
  \begin{itemize}
  \item Description: rough surfaces that result in diffuse reflection
  \item Example: dry asphalt surface, building, bridge pillar
  \item Detection difficulty: medium
  \end{itemize}
\item{\bf Retroreflective surface}
  \begin{itemize}
  \item Description: a surface that reflects light back to its original direction with minimum scattering
  \item Example: safety clothing, traffic sign
  \item Detection difficulty: easy
  \end{itemize}
\end{itemize}

Specular surfaces are the most challenging to detect due to how light is reflected \cite{Lichti2002THEEO}. Some LiDAR models have a certain level of self-correction and -calibration capabilities, but in most cases, preprocessing steps need to be carried out before feeding the LiDAR measurements to the downstream algorithms. Retroreflective surfaces are typically the easiest to detect, but one should bear in mind that there is a discrepancy between the theoretical properties and the practical sensor output, which depends on the design of the LiDAR instrument as well.

Cautious measures need to be taken in order to make sure that these influencing factors do not have a critical impact on the system output. Careful examination of the LiDAR design manual and thorough analysis of the produced point cloud is a necessity to understand the theoretical limitation and the data quality in practice. Moreover, preprocessing steps, such as outlier rejection and intensity calibration \cite{CK17}, are usually needed before feeding the LiDAR data to the detection algorithms. Outlier rejection refers to the preprocessing step associated with the measured 3D position and intensity calibration is the adjustment of the intensity given the influencing factors. In \cite{kashani2015review}, the authors have reviewed both rigorous and ad hoc calibration techniques. Rigorous calibration is further categorized into four levels: no modification, intensity correction, intensity normalization and radiometric correction. Ad hoc calibration is leaning towards a more data-driven fashion. The choice of the exact calibration method largely depends on the use case. It is necessary to evaluate the output from the detection system as a whole to assess the importance of these influences in order to choose the least complex yet sufficient preprocessing techniques.

\paragraph{Annotation} The LiDAR sensor is often considered as the reference sensor due to the fact that it accurately (up to centimeter spatial resolution) captures the 3D information of the environment, which gives the possibility of extracting the geometric features of the 3D world. Typical LiDAR annotations for roadside object detection include 3D bounding boxes, pointwise panoptic segmentation and pointwise semantic segmentation.

Similar to images, these annotations can be either manually labelled or produced by an automated process. One advantage of a LiDAR is that automation for annotation is made possible to a large extent by the 3D information it provides. Moreover, when a LiDAR sensor is synchronized to video cameras, it is also possible to automatically annotate the camera images using the LiDAR information. This automation enables better scalability and potentially higher consistency. 

\paragraph{Algorithms} There are three most important categories of algorithms when it comes to LiDAR data processing: 1) geometric analysis, 2) radiometric processing and 3) deep learning algorithms, where 1) and 2) are considered model-based methods, whereas 3) is data-driven.

Geometric analysis \cite{wang2017sigvox, poux2019voxel} imposes certain assumptions on the shape of the object. For instance, street poles close to the road are one of the most important hazardous objects that need to be detected. They can be extracted largely based on their distinct geometric properties \cite{li2019localization}. These methods are typically unsupervised, which reduces the time and cost for producing annotations.

Radiometric processing concerns the analysis of the reflected light intensity. As mentioned above, there are multiple influencing factors that affect the reflected intensity, such as the material of the detected surface and data acquisition geometries. For roadside object detection, together with the geometric information, these influencing factors can be used for calibration and detecting certain surfaces \cite{Misener1998InvestigationOA, shin2019characteristics}.

To further improve the detection and analysis of the objects, deep learning techniques for LiDAR processing can be adopted. However, due to the high demand of expensive computational resources, such as high-end GPU servers and annotations, deep learning algorithms are typically applied as a complementary step after sophisticated model-based data processing. This is an effective approach for LiDAR data because of the rich and structured information contained in the measurement.

\paragraph{Limitations} The range of a reasonably priced LiDAR sensor can be a limiting factor for algorithm development. For example, the 32 layer Velodyne LiDAR sensor is a commonly used device for perception systems on the road and its range is less than 100 meters. Fortunately, this is usually not an issue for offline detection of static roadside objects since multiple scans can be registered to create a static map as the data collection vehicle moves forward.

Multipath reflection is another potential issue, where a laser beam hits multiple objects before returning to the receiver. This effect can be reduced by applying outlier detection techniques. Furthermore, when combining a LiDAR point cloud with camera images, another potential risk is the accuracy of the time synchronization. This problem can be improved by designing a logging system where different sensors are triggered at the same time. If modifying the logging system is not an option, time compensation needs to be applied before the sensor fusion step.

\subsubsection{Summary}
A typical roadside object detection system consists of one or multiple time synchronized LiDAR scanners and cameras, where different sensors are used to complement each other and enrich the system perception as a whole. These detection systems are almost always equipped with GNSS receivers and IMU sensors to utilize the geometric information and ego motion. The basic processing of the GNSS data is rather standardized. Nevertheless, more advanced geographical information processing might provide better context of the data collection environment and further improve the detection results \cite{SILVA2017415}.

Due to its capability of building 3D models and the high spatial resolution, the LiDAR scanner is often used as the reference sensor in an object detection system. The choice of the exact instruments and algorithms depends on the use case and the requirements of the detection output. Camera sensors have relatively low cost with high technical maturity. They are often used in real-time systems for pattern recognition. There is extensive research and development on the subject of object classification and segmentation using image-based deep learning algorithms. Automated annotation and transfer learning techniques can be adopted to reduce the burden of manual annotation, where the automation typically relies on the LiDAR sensor.

\subsection{Commercial Solutions}
Most of the state-of-the-art technologies described in \ref{sec:sot} are available in the form of open resources. Assembling the whole system requires specialized resources, expertise and experiences. Therefore, some users prefer to outsource this activity.
The most commonly available commercial products on the market are the ones that cover the entire pipeline from the data collection platform to detection results. These solutions offer an end-to-end system, which have the benefit of convenience with the potential drawback of being expensive and inflexible.
An alternative is to separate the data collection activity from the data processing to find suitable providers that are specialized in each of these domains. This solution typically brings better flexibility, whereas the communication and collaboration between the data collection and data processing may be more challenging.

\section{Available Data Sources}

In this section, we investigate two potential data sources for roadside object detection:
\begin{itemize}
\item[1)] Surveying vehicle: annual scans from the Swedish Transport Administration.
\item[2)] Consumer vehicle: data collected using a research data collection vehicle.
\end{itemize}
The first setup is designated for road surveying, whereas the second is equipped with a typical consumer-oriented AD/ADAS data collection and perception system.
In practice, both setups are commonly used for roadside object detection and hence we cover both of them in this section, where the sensors and their specifications are described to provide a better understanding of the capabilities and limitations of the systems, as well as to estimate what is needed to scale up the process.

\subsection{Annual Scans from Trafikverket}
\begin{figure}[h!]
  \centering
  \begin{subfigure}[b]{0.3\textwidth}
    \includegraphics[width=\textwidth]{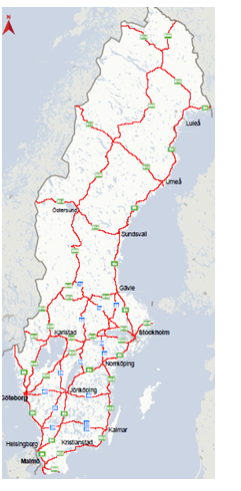}
  \end{subfigure}
  \caption{Illustrative map of the Trafikverket annual scanning data. The red lines in the image are showing the data coverage.}
  \alt{Major roads in Sweden that are covered by the Trafikverket annual scan.}
  \label{fig:scan_map}
\end{figure}

One data source is the annual continuous scan of the entire paved state road network (excluding gravel roads). The data is collected by a fleet of data collection vehicles. One example is the scanning from 2016.
The size of the collected data amounts to hundreds of TB today. The image in Fig.~\ref{fig:scan_map} shows the major roads that have been scanned in this data collection project.

\subsubsection{Data collection setup}

\paragraph{Camera image}

\begin{figure}[h!]
    \centering
    \begin{subfigure}[b]{0.30\textwidth}
        \includegraphics[width=\textwidth]{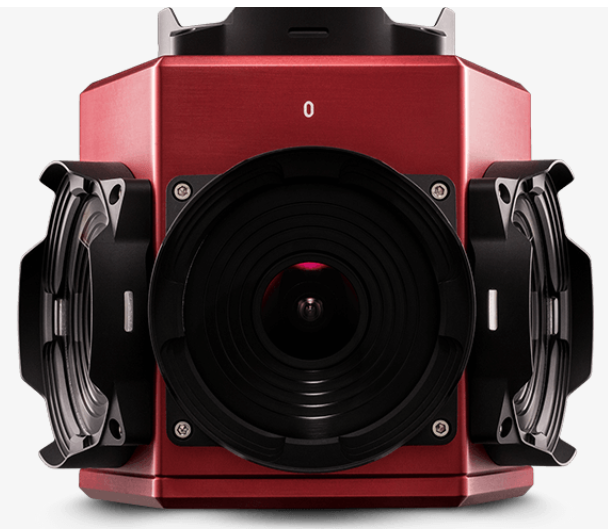}
        \label{fig:trafikverket_camera_device}
    \caption{The Ladybug5 surround vision camera.}
  \end{subfigure}
  ~~~~~~~
    \centering
    \begin{subfigure}[b]{0.60\textwidth}
        \includegraphics[width=\textwidth]{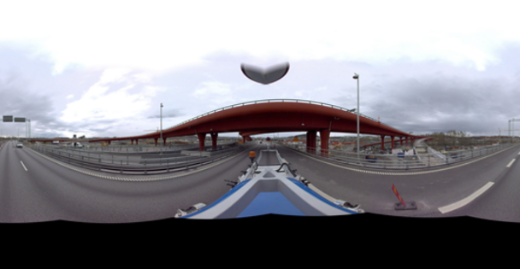}
    \caption{Example image produced by the camera to the left.}
  \end{subfigure}
  \caption{Example data collection camera for the Trafikverket annual scans.}\label{fig:camera}
  \alt{Example picture of the fisheye camera and its output image.}
        \label{fig:trafikverket_camera}
\end{figure}

The camera device is required to produce images that have a color depth of at least 12 bits. The pictures must be comprehensive in both driving directions. The image resolution should be at least 30 megapixel. One example camera is shown in Fig.~\ref{fig:camera}. It is the 360 degrees Ladybug5 with the following specifications.
\begin{itemize}
\item Format: JPG
\item File size: 6-9 MB
\item Resolution: 32 megapixel
\item Color depth: 12 bit
\end{itemize}

\begin{figure}[th!]
    \centering
    \begin{subfigure}[b]{0.28\textwidth}
        \includegraphics[width=\textwidth]{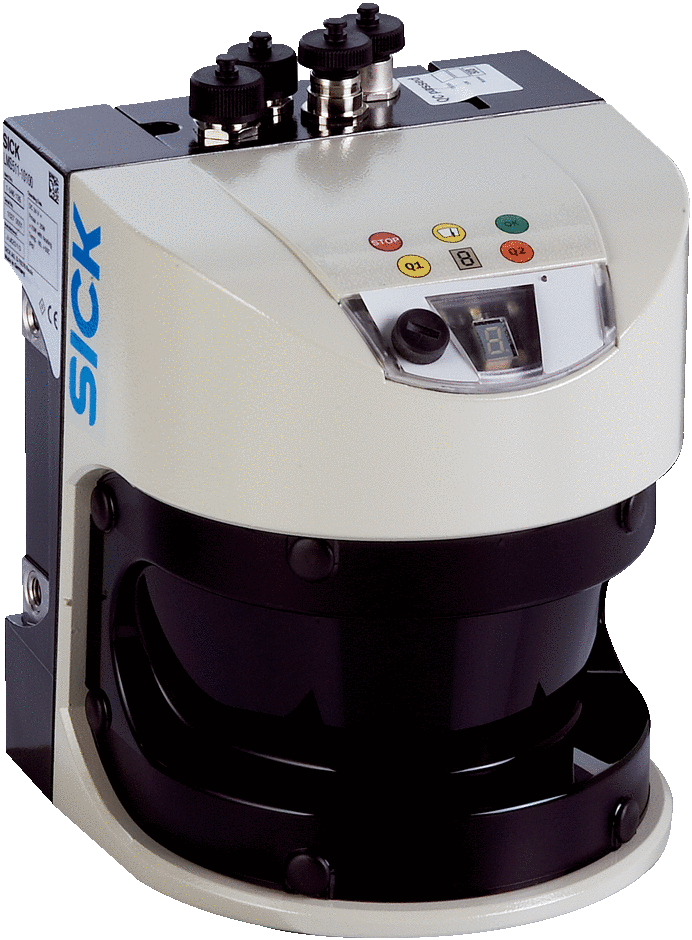}
        \label{fig:trafikverket_lidar_device}
    \caption{The SICK LMS511 PRO LiDAR scanner.}
  \end{subfigure}
  ~~~~~~~~~~
    \centering
    \begin{subfigure}[b]{0.60\textwidth}
        \includegraphics[width=\textwidth]{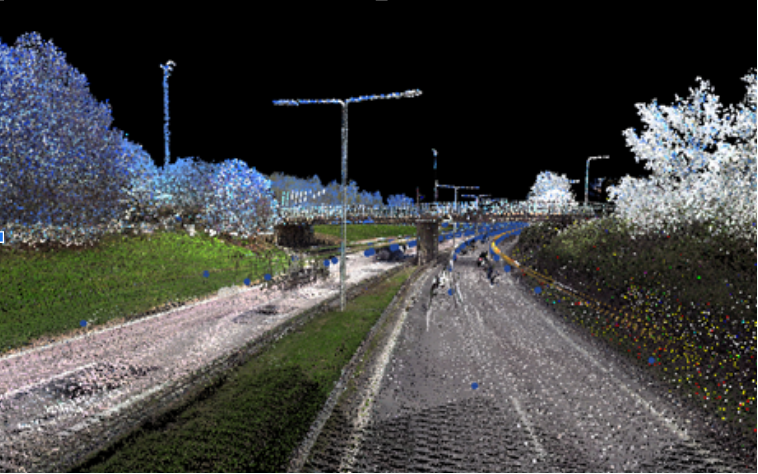}
    \caption{Example point cloud produced by the LiDAR scanner.}
    \end{subfigure}
    \caption{Example data collection LiDAR scanner for the Trafikverket annual scans.}
    \alt{Example LiDAR device.}
        \label{fig:trafikverket_lidar}
\end{figure}

\paragraph{LiDAR Point cloud}
At least two scanners should be used, rotated 135 degrees in relation to the direction of travel, to minimize shadow effects.
One example of the LiDAR scanners is the SICK LMS511 PRO shown in Fig.~\ref{fig:trafikverket_lidar},
which produces a point density of about 1500 points per meter square on the road surface and is usually collected once in each direction of the road (applies directly under the vehicle and decreases outwards).

\paragraph{GNSS/IMU platform}
One example device for global positioning and inertial navigation is the OxTS land surveyor inertial+ platform. It provides the following measurements and estimates:
\begin{itemize}
\item Position and velocity
\item Wheel speed
\item Angular rate
\item Heading, pitch, roll
\item Acceleration
\end{itemize}
\begin{figure}[h!]
    \centering
    \begin{subfigure}[b]{0.40\textwidth}
        \includegraphics[width=\textwidth]{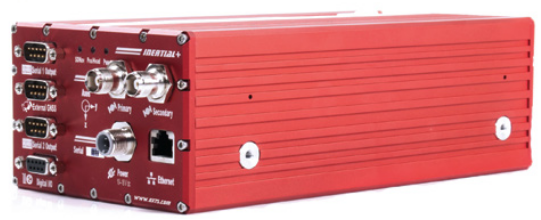}
  \end{subfigure}
  \caption{The OxTS land surveyor inertial+ platform.}
  \alt{OxTS land surveyor inertial+.}
  \label{fig:trafikverket_gnss_device}
\end{figure}

\subsubsection{Data size}
\begin{figure}[h!]
    \centering
    \begin{subfigure}[b]{0.6\textwidth}
        \includegraphics[width=\textwidth]{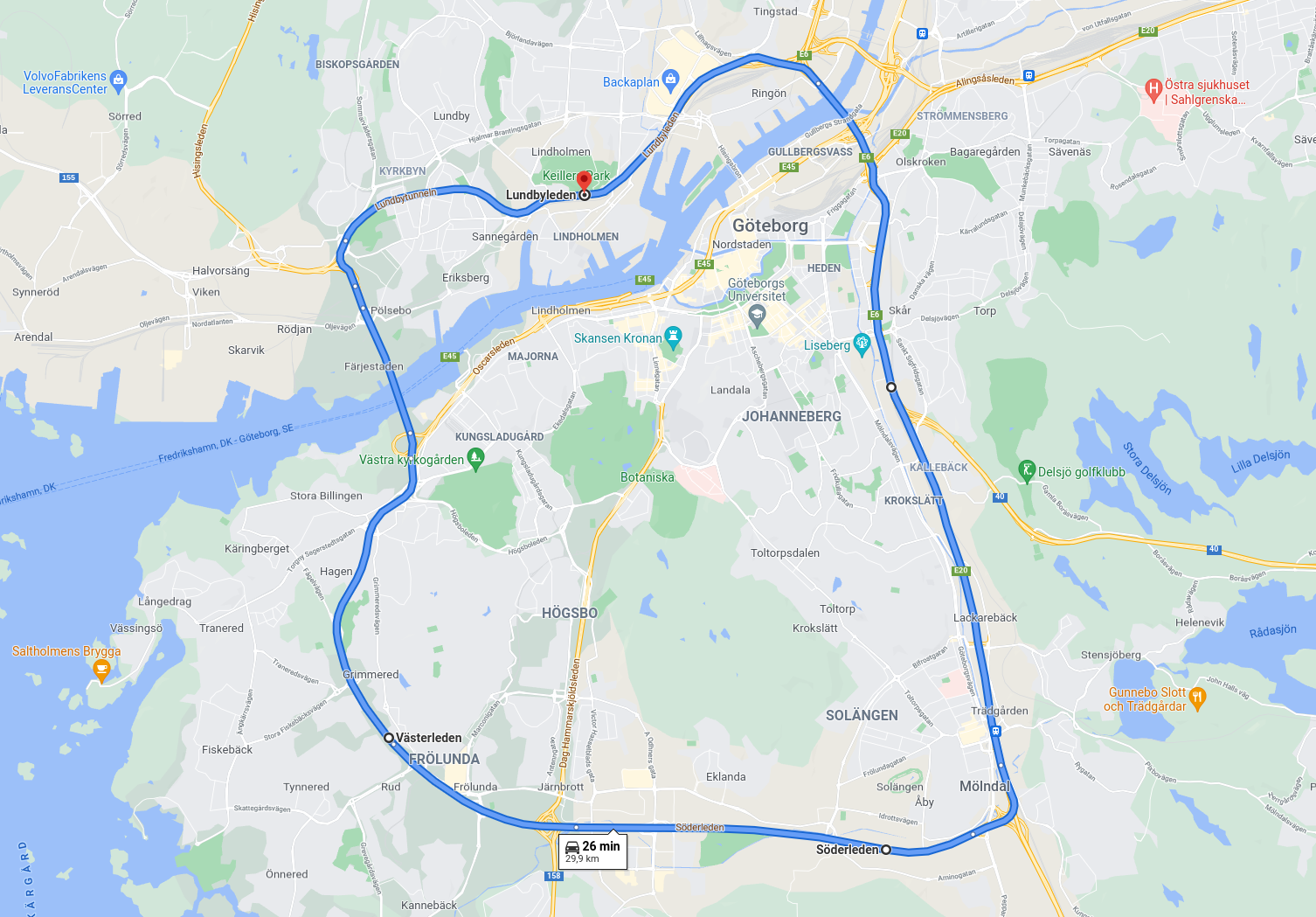}
    \end{subfigure}
    \caption{One lap around Slingan. Sligan is a round trip that has been historically used as a test track for Volvo Cars in the city of G\"{o}teborg.}
    \alt{A map showing Slingan.}
    \label{fig:slingan}
\end{figure}

As described above, the scanning data from Trafikverket consists of point clouds, camera images and GNSS/IMU data. The image data collected by the 360 degrees surround camera resulting in around 6-9 MB JPEG image each meter of the road or 140 MB/s at 70 km/h.
The point cloud data is collected from multiple LiDAR sensors. The number of sensors may vary. For the purpose of estimating the scalability, we take an example set up with eight LiDAR sensors, which results in a point density of 1500 points/$m^2$ on the road surface below the collection vehicle. This would result in approximately 10000 points per driven meter or 3.2 MB/s at 70 km/h. An estimate of the data size can be found
in Table~\ref{tab:trafikverket_data}.
One lap around Slingan (about 30 km for 26 minutes illustrated in Fig.~\ref{fig:slingan}) would produce a data size of about 215 GB.

\begin{table}[h!]
  \begin{center}
    \resizebox{\columnwidth}{!}{
    \begin{tabular}{| c | c | c | c | c | }
    \hline
    Type & Quantity & Frequency & Bandwidth & Total \\\hline
    Camera, 32MP & 1 & 1/meter & 140 MB/s & 140 MB/s \\\hline
    LiDAR  & 8 & 1500 pts/$m^2$ & 0.4 MB/s & 3.2 MB/s \\\hline
    GNSS/IMU  & 1 &  100 or 250 Hz  & 1 KB/s & 1 KB/s \\\hline
      {\bf Total}  & - & - & - & 143 MB/s \\\hline\omit
    \end{tabular}
    }
    \end{center}
    \caption{Example size of Trafikverket data collection.}
    \label{tab:trafikverket_data}
\end{table}

\subsection{Data collection using the REVERE research vehicle}

For collecting reference data from specified road areas for development purposes, we have used the Revere Snowfox vehicle. It is equipped with a set of sensors, including cameras, lidar, RTK GPS, and a logging system to record raw data, timestamping etc.

\begin{table}[h!]
  \centering
    \resizebox{\columnwidth}{!}{
    \begin{tabular}{| c | c | c | c | c |}
    \hline
    Type & Quantity & Frequency & Bandwidth & Total \\\hline
    Camera, 3.1MP & 5 & 16 Hz & 33 MB/s & 165 MB/s \\\hline
    Camera, 0.92MP & 1 & 25 Hz & 1 MB/s & 1 MB/s \\\hline
    LiDAR, 32 layer  & 1 & 10 Hz & 2 MB/s & 2 MB/s \\\hline
    RTK GPS  & 1 & 20 Hz & 1 KB/s & 1 KB/s \\\hline
    {\bf Total}  & - &  -& - & 168 MB/s \\\hline\omit
    \end{tabular}
    }
    \caption{Example data size of data collection using the REVERE research vehicle.}
    \label{tab:revere_data}
\end{table}

As shown in Table~\ref{tab:revere_data}, the total data produced is approximately 168 MB/s. One lap around “slingan” (30 km/25 minutes) would produce 252 GB data. To achieve robust results from data driven algorithms, it is important to have a large variety in the training and validation data sets. Examples of such varieties include weather conditions, lighting conditions and various traffic densities. To achieve a high variety, 10-20 collections in the same area are required. This would result in 5 TB of data storage in total.

\section{Estimated resources}

\subsection{Data storage hardware}
There are several parameters that will set the requirements on storage servers. In addition to the storage space there are parameters specifying transfer speeds and latency.
Generally speaking, the storage consist of four parts:
\begin{itemize}
\item Storage for the collected raw data
\item Annotations and algorithm results
\item Development and deployment environment
\item Data caching
\end{itemize}
A common way to deploy algorithms is by using a containerized environment hosted on virtual machines. A virtual machine typically uses around 50 GB storage. An application container image often consumes around  10 GB. In a typical project of this nature, we estimate that around 10 such applications will be developed and deployed.
This is summarized in Table~\ref{tab:storage}.
The algorithm output is estimated to require a storage space of up to 50\% of the raw data. The storage required for data caching depends on the data pipeline, algorithm used and other factors and is therefore not included in the table.

\begin{table}[h!]
  \centering
    \begin{tabular}{| c | c | c | c |}
    \hline
    Type of data& Quantity & Size &  Total \\\hline
    Raw data & - & 5 TB & 5 TB \\\hline
    Annotations and algorithm outputs & - &2.5 TB&2.5 TB \\\hline
      Virtual machines & 3 & 50 GB & 150 GB  \\\hline
      Applications & 10 & 10 GB & 100 GB \\\hline
      {\bf Total} & - & - & 7.75 TB \\\hline\omit
    \end{tabular}
    \caption{Data storage requirement}
    \label{tab:storage}
\end{table}

\subsection{Deep learning models}

Deep learning models are typically trained and running on GPUs for high computational speed and efficiency. There are specific requirements on the CPU, RAM and storage.
\subsubsection{Training}
\label{sec:training}
Training a deep learning model using 10 000 data samples would correspond to a dataset of a size around 150 GB.
When training a model on a GPU, the GPU memory is often a hard limiting factor. To train a model of this size, it would require at least 24 GB with support for 16 bit floating point.

Typically, when training on one GPU, the training speed is around 5 data samples per second, which results in around 75 MB/s. Based on this assumption, training this model for 50 epochs would take around 28 hours. The time for preprocessing and augmenting one sample is estimated to be 1 second per CPU core with 4 GB of RAM memory. Given this estimate, 5 CPU cores per GPU are required to be able to fully saturate the GPU resources. To reduce the turnaround time to train a model, multiple GPUs can be used. The computation speedup is expected to be close to linear with the number of GPUs. Using a compute node with 4 GPUs would result in a training time of 7 hours, requiring nodes with 20 CPU cores and 80 GB RAM, where 300 MB/s data is consumed.

To satisfy the bandwidth requirements and highly random read patterns for deep learning training, it is recommended to use solid state drives (SSD) connected with at least 2.5 Gbit/s data interface for a 4 GPU node.

Given these estimates and the assumption that different versions of a model need to be trained 30 times, it would result in 840 GPU hours. If a 4 GPU node is used, the total training time is expected to be 210 hours (see ~Table \ref{tab:training}).
\begin{table}[h!]
  \centering
    \begin{tabular}{| c | c | c | }
    \hline
      Resource &  Specifications & Estimated amount \\\hline
     \multirow{2}{*}{Compute}  & 20 cores, 80 GB mem, 4 GPUs &\multirow{2}{*}{210 hours}\\
      &  24 GB VRAM, FP16 & \\\hline
      Storage & 300 MB/s SSD  & 150 GB \\\hline\omit
    \end{tabular}
    \caption{Deep learning training}
    \label{tab:training}
\end{table}

\subsubsection{Inference}
Deep learning inference is assumed to take around half of the time compared to training, i.e., around 10 samples per second per GPU. Preprocessing for inference is often lighter compared to training. The compute requirements for preprocessing during inference is assumed to be half of what is required for training, resulting in 3 cores per GPU and 2 GB per CPU core. Given the collected data described above, it would take around 30 minutes to process one round around ``Slingan''. Given 10 different collections, this would add up to 5 hours on one GPU. Assuming 30 different models, this would result in 150 GPU hours, or 37.5 hours on a 4 GPU node. A 4 GPU node would require a storage bandwidth of 600 MB/s (see ~Table \ref{tab:inference}).

\begin{table}[h!]
  \centering
    \begin{tabular}{| c | c | c | }
    \hline
      Resource &  Specifications & Estimated amount \\\hline
     \multirow{2}{*}{Compute}  & 12 cores, 24 GB mem, 4 GPUs &\multirow{2}{*}{37.5 hours}\\
      &  24 GB VRAM, FP16 & \\\hline
      Storage & 600 MB/s SSD  & 167 GB \\\hline\omit
    \end{tabular}
    \caption{Deep learning inference}
    \label{tab:inference}
\end{table}

\subsection{Scale up}
The distance covered by the entire data collection from Trafikverket is about 98400 km \cite{trafikverket_data}, which is 3000 times longer than Sligan.
To scale up the development, there are four main considerations:
\begin{itemize}
\item Additional storage: It is obvious that additional storage capacity needs to be added.
  The storage does not need to be expensive fast SSD storage as required for training since the performance requirement for data traversal is more relaxed.
  However, fast storage can boost the speed of the inference, where smart data management and scheduling functionalities can be enabled to optimize the balance between cost and efficiency.
\item Additional training: Depending on the complexity of data and the requirement of the inference accuracy, additional training may be needed to certain extents, which is then translated into more training time and computational resources.
  The variation of this estimate depends on many factors.
  We loosely estimate the overall training data required for static roadside object detection to be around 10 times the initial training step estimated in \ref{sec:training}.
\item Inference: The resources allocated for inference does not need to scale with the data size if there are no strict requirements on the inference time. If no additional resources are introduced, the inference time will grow linearly with respect to the size of the data collection. The results from the inference are not necessarily stored depending on the workflow and use case.
\item Validation: The detection results produced by the system need to be tested and validated, which means that some of the results, such as anomalies and common mistakes, need to be stored, analyzed and used as feedback for additional training steps. Moreover, to be able to validate the detection system, a large amount of data need to be annotated, where manual annotation may not be scalable. These annotations need to be stored as well. It is not required to scale up the fast SSD storage for this purpose. The exact amount depends on the level of trust required for the system.
\end{itemize}

\section{Conclusion}

In this document, we investigate how to implement and scale up a roadside object detection system in order to design, manage and maintain safer road infrastructure.
We describe a data processing system from three aspects: 1. the target of interest, 2. the sensors of choice for data collection and 3. the algorithms applied for object detection and classification.

The target of interest is mainly derived from two sources: 1) a list of hazardous roadside objects from iRAP \cite{irapcodingmanual}, the International Road Assessment Programme, and 2) a list of important roadside objects from Trafikverket, the Swedish Transport Administration.
We conclude that cliffs are the most critical to detect, which is followed by rigid point objects, such as poles and trees, close to the road and more than 10cm in diameter. Continuous objects, such as slopes, vertical faces and ditches, are crucial to detect as well. The presence of such hazardous objects need to be identified with a low type II error (false negative). On the other hand, protective objects, such as crash cushions and guard rails need to be detected with a low type I error (false positive).

Given these objects of interest, various sensor setups for data collection are discussed. A typical setup includes LiDAR scanners, cameras and GNSS/IMU devices. Cameras are commonly used for object detection and classification. They work the best when the 2D characteristics, such as contours and colors, are well represented. However, for the task of roadside object detection, the 3D structure plays an important role in order to measure distances, shapes and volumes. In these scenarios, a well designed LiDAR scanner may provide the necessary information with a high precision. Due to this reason, LiDAR is often used as a reference sensor to automate the annotation process for camera images. The downside of a LiDAR device is that it can be expensive and sensitive to certain weather conditions, target materials and geometries. Moreover, the data processing procedure can be complex in order to extract the structural information of objects.
These sensors are almost always accompanied by a GNSS/IMU platform for global time stamping and georeferencing.

For both 2D and 3D object detection and classification, model-based algorithms and data-driven techniques can be applied. Model-based algorithms are in general more analytical and interpretable with the risk of being inaccurate in its model assumptions, which may cause a large bias. On the other hand, data-driven techniques, such as deep learning, may exhibit exceptional performance on certain tasks with the downside of being unpredictable and demanding on the high varieties of training data and annotations. One way forward is therefore to combine the strengths of both techniques to build a robust system.

To estimate the resources needed for a large-scale road scanning and roadside object detection system, we investigate two major data sources: 1) annual scans from Trafikverket and 2) data collected using a research vehicle from REVERE, the laboratory of resource for vehicle research at Chalmers, which represent a road surveying setup and a consumer's vehicle setup, respectively. We have provided an estimate of the data storage and compute resources needed for developing and deploying such a system as well as an indication of how it scales with respect to the growth of data collection.

\section{Summary}
In this document, we investigate roadside object detection systems including data collection and processing pipelines for both development and deployment with the objective to provide a better understanding on how to design, implement and scale up such a system.
A well managed and maintained road network is the key to economic growth and traffic safety. It is Trafikverket's vision to ensure that ``everybody arrives smoothly, the green and safe way''.
The system studied in this document is one of the enablers towards safer road infrastructure and Sweden's Vision Zero.

\section{Acknowledgement}
This work is a collaboration between SAFER, Trafikverket, Asymptotic AI and Volvo Cars. It is financed by the SAFER pre-study program (2021).

The images used for illustrating data analysis are part of the preliminary results from the project ``Safety-driven data labelling platform to enable safe and responsible AI'' (2020-02952). The data is collected using the REVERE research vehicle. This work is financed by Vinnova and implemented by Asymptotic AI.

\bibliography{refs}
\bibliographystyle{abbrv}
\end{document}